\documentclass{article}
\usepackage{ijcai19}

\usepackage{times}
\usepackage{soul}
\usepackage{url}
\usepackage[hidelinks]{hyperref}
\usepackage[utf8]{inputenc}
\usepackage[small]{caption}
\usepackage{graphicx}
\usepackage{amsmath}
\usepackage{booktabs}
\title{Image-to-Image Translation with Multi-Path Consistency Regularization}

\author{
Jianxin Lin$^1$\thanks{The first two authors contributed equally to this work}\and
Yingce Xia$^3$\footnotemark[1]\and
Yijun Wang$^2$\and
Tao Qin$^{3}$\And
Zhibo Chen$^1$\footnote{Corresponding author}\\
\affiliations
$^1$CAS Key Laboratory of Technology in Geo-spatial Information Processing and Application System, University of Science and Technology of China \\
$^2$University of Science and Technology of China\\
$^3$Microsoft Research Asia\\
\emails
\{linjx, wyjun\}@mail.ustc.edu.cn,
\{yingce.xia, taoqin\}@microsoft.com,
chenzhibo@ustc.edu.cn
}

\begin{document}

\maketitle

\begin{abstract}
   Image translation across different domains has attracted much attention in both machine learning and computer vision communities. Taking the translation from source domain $\mathcal{D}_s$ to target domain $\mathcal{D}_t$ as an example, existing algorithms mainly rely on two kinds of loss for training: One is the discrimination loss, which is used to differentiate images generated by the models and natural images; the other is the reconstruction loss, which measures the difference between an original image and the reconstructed version through $\mathcal{D}_s\to\mathcal{D}_t\to\mathcal{D}_s$ translation. In this work, we introduce a new kind of loss, multi-path consistency loss, which evaluates the differences between direct translation $\mathcal{D}_s\to\mathcal{D}_t$ and indirect translation $\mathcal{D}_s\to\mathcal{D}_a\to\mathcal{D}_t$ with $\mathcal{D}_a$ as an auxiliary domain, to regularize training. For multi-domain translation (at least, three) which focuses on building translation models between any two domains, at each training iteration, we randomly select three domains, set them respectively as the source, auxiliary and target domains, build the multi-path consistency loss and optimize the network. For two-domain translation, we need to introduce an additional auxiliary domain and construct the multi-path consistency loss. We conduct various experiments to demonstrate the effectiveness of our proposed methods, including face-to-face translation, paint-to-photo translation, and de-raining/de-noising translation.
\end{abstract}
\section{Introduction}
\begin{figure}[!ht]
\centering
\includegraphics[width=0.95\linewidth]{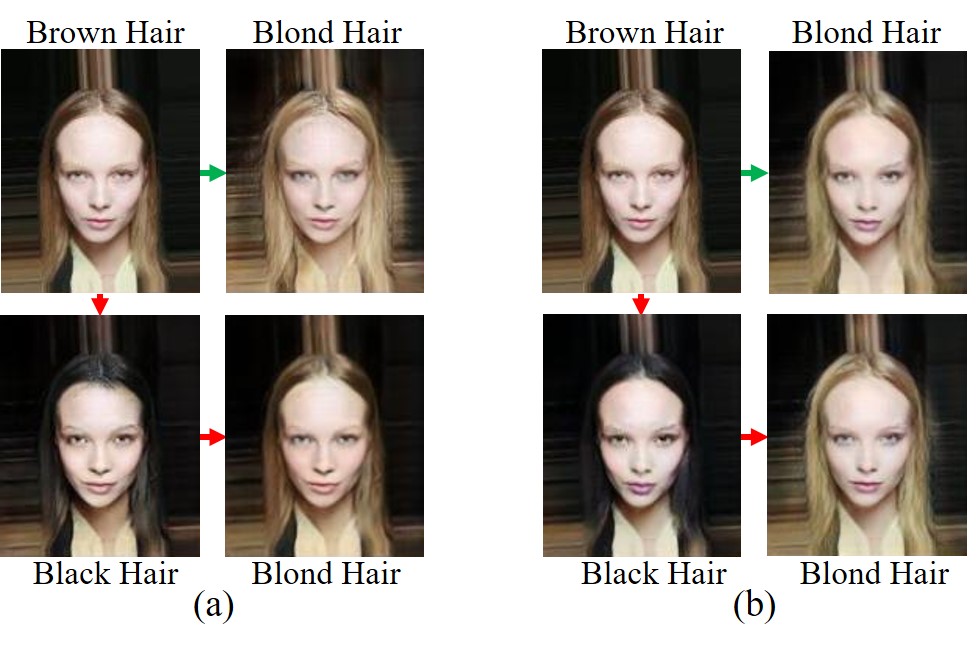}
\caption{Illustration of multi-path consistency regularization on the translation between different hair colors. (a) Results of StarGAN. (b) Our results.}
\label{fig:illustration}
\end{figure}
Image-to-image translation aims at learning a mapping that can transfer an image from a source domain to a target one, while maintaining the main representations of the input image from the source domain. Many computer vision problems can be viewed as image-to-image translation tasks, including image stylization~\cite{gatys2016image}, image restoration~\cite{mao2016imageResto}, image segmentation~\cite{girshick2015fast} and so on. Since a large amount of parallel data is costly to collect in practice, most of recent works have focused on unsupervised image-to-image translation algorithms. Two kinds of algorithms are widely adopted for this problem. The first one is generative adversarial networks (briefly, GAN), which consists of an image generator used to produce images and an image discriminator used to verify whether an image is a fake one from a machine or a natural one. Ideally, the training will reach an equilibrium where the generator should generate ``real'' images that the discriminator cannot distinguish from natural images~\cite{goodfellow2014generative}. The other one is dual learning~\cite{he2016dual}, which is first proposed for neural machine translation and then successfully adapted into image translation. In dual learning, a pair of dual tasks is involved, like man-to-woman translation v.s. woman-to-man translation. The reconstruction loss of the two tasks are minimized during optimization. The combination of GAN algorithms and dual learning leads to many algorithms for two-domain image translation like CycleGAN~\cite{Zhu_2017_ICCV}, DualGAN~\cite{Yi_2017_ICCV}, DiscoGAN~\cite{kim2017learning}, conditional DualGAN~\cite{lin2018conditional}, and for multi-domain image translation like StarGAN~\cite{choi2017stargan}.

We know that multiple domains are bridged by multi-path consistency. Take the pictures in Figure~\ref{fig:illustration} as an example. We want to work on a three-domain image translation problem, which targets at changing the hair color of the input image to a specific one. Ideally, the direct translation (i.e., one-hop translation) from brown hair to blond should be the same as the indirect translation (i.e., two-hop translation) from brown to black to blond. However, such an important property is ignored in current literature. As shown in Figure~\ref{fig:illustration}(a), without multi-path consistency regularization, the direct translation and indirect translation are not consist in terms of hair color. Besides, on the right of the face in the one-hop translation, there is much horizontal noise. To keep the two generated images consistent, in this paper, we propose a new loss, {\em multi-path consistency loss}, which explicitly models the relationship among three domains. We require that the differences between direct translation from source to target domain and indirect translation from source to auxiliary to target domain should be minimized. For example, in Figure~\ref{fig:illustration}, the $L_1$-norm loss of the two translated blond hair pictures should be minimized. After applying this constraint, as shown in Figure~\ref{fig:illustration}(b), the direct and indirect translations are much similar, and both the direct and indirect translations are of less noise.

Multi-path consistency loss can be generally applied in image translation tasks. For multi-domain ($\ge3$) translation, during each training iteration, we can randomly select three domains, apply the multi-path consistency loss to each translation task, and eventually obtain models that can generate better images. For the two-domain image translation problem, we need to introduce a third auxiliary domain to help establish the multi-path consistency relation.

Our contributions can be summarized as follows: (1) We propose a new learning framework with multi-path consistency loss that can leverage the information among multiple domains. Such a loss function regularizes the training of each task and leads to better performance. We provide an efficient algorithm to optimize such a framework. (2) We conduct rich experiments to verify the proposed method. Specifically, we work on face-to-face translation, paint-to-photo translation, and de-raining/de-noising translation. For qualitative analysis, the models after applying multi-path consistence loss can generate clearer images with less blocks artifacts. For quantitative analysis, we calculate the classification errors and PSNR for tasks, which all outperform the baselines. We also conduct user study on the multi-domain translation task and $59.14\%$/$89.85\%$ users vote for that our proposed method is better on face-to-face and paint-to-photo translations.


\section{Related works}\label{related work}
In this section, we summarize the literature about GAN and unsupervised image-to-image translation.

\noindent\textbf{GAN} GAN~\cite{goodfellow2014generative} was firstly proposed to generate images in an unsupervised manner. A GAN is made up of a generator and a discriminator. The generator maps a random noise to an image and the discriminator verifies whether the image is a natural one or a fake one. The training of GAN is formulated as a two-player minmax game. Various versions of GAN have been proposed to exploit its capability for different image generation tasks \cite{arjovsky2017wasserstein,huang2017stacked,lin2018multiscale}. InfoGAN~\cite{chen2016infogan} learns to disentangle latent representations by maximizing the mutual information between a small subset of the latent variables and the observation. \cite{radford2015unsupervised} presented a series of deep convolutional generative networks (DCGANs) for high-quality image generation and unsupervised image classification tasks, which bridges the convolutional neural networks and unsupervised image generation together. SRGAN \cite{ledig2016photo} maps low-resolution images to high resolution images. Isola et al.~\cite{isola2016image} proposed a general conditional GAN for image-to-image translation tasks, which could be used to solve label-to-street scene and aerial-to-map translation problems.


\noindent\textbf{Unsupervised image-to-image translation} Since it is usually hard to collect a large amount of parallel data for supervised image-to-image translation tasks, unsupervised learning based algorithms have been widely adopted. Based on adversarial training, Dumoulin et al. \cite{dumoulin2016adversarially} and Donahue et al. \cite{donahue2016adversarial} proposed algorithms to jointly learn mappings between the latent space and data space bidirectionally. Taigman et al. \cite{taigman2016unsupervised} presented a domain transfer network (DTN) for unsupervised cross-domain image generation under the assumption that a constant latent space between two domains exits, which could generate images of target domain's style and preserve their identity. Inspired by the idea of dual learning~\cite{he2016dual}, DualGAN~\cite{Yi_2017_ICCV}, DiscoGAN~\cite{kim2017learning} and CycleGAN~\cite{Zhu_2017_ICCV} were proposed to tackle the unpaired image translation problem by jointly training two cross-domain translation models. Meanwhile, several works~\cite{choi2017stargan,liu2018unified} have been further proposed for multiple domain image-to-image translation with a single model only. 



\section{Framework}\label{framework}
In this section, we introduce our proposed  framework built on multi-path consistency loss. Suppose we have $N$ different image domains $\{\mathcal{D}_1, \mathcal{D}_2,..., \mathcal{D}_N\}$ where $N\ge2$. A domain can be seen as a collection of images. Generally, the image translation task aims at learning $N(N-1)$ mappings $f_{i,j}:\mathcal{D}_i\mapsto\mathcal{D}_j$ where $i\ne j$. Also, we might come across the cases that we are interested in a subset of the $N(N-1)$ mappings. We first show how to build translation models between $\mathcal{D}_i$ and $\mathcal{D}_j$ with $\mathcal{D}_k$ as an auxiliary domain and then present the general framework for multi-domain image translation with consistency loss. Note that in our framework, $i,j,k\in[N]$ and they are different to each other.
\begin{figure*}[ht!]
\centering
\includegraphics[width=0.9\linewidth]{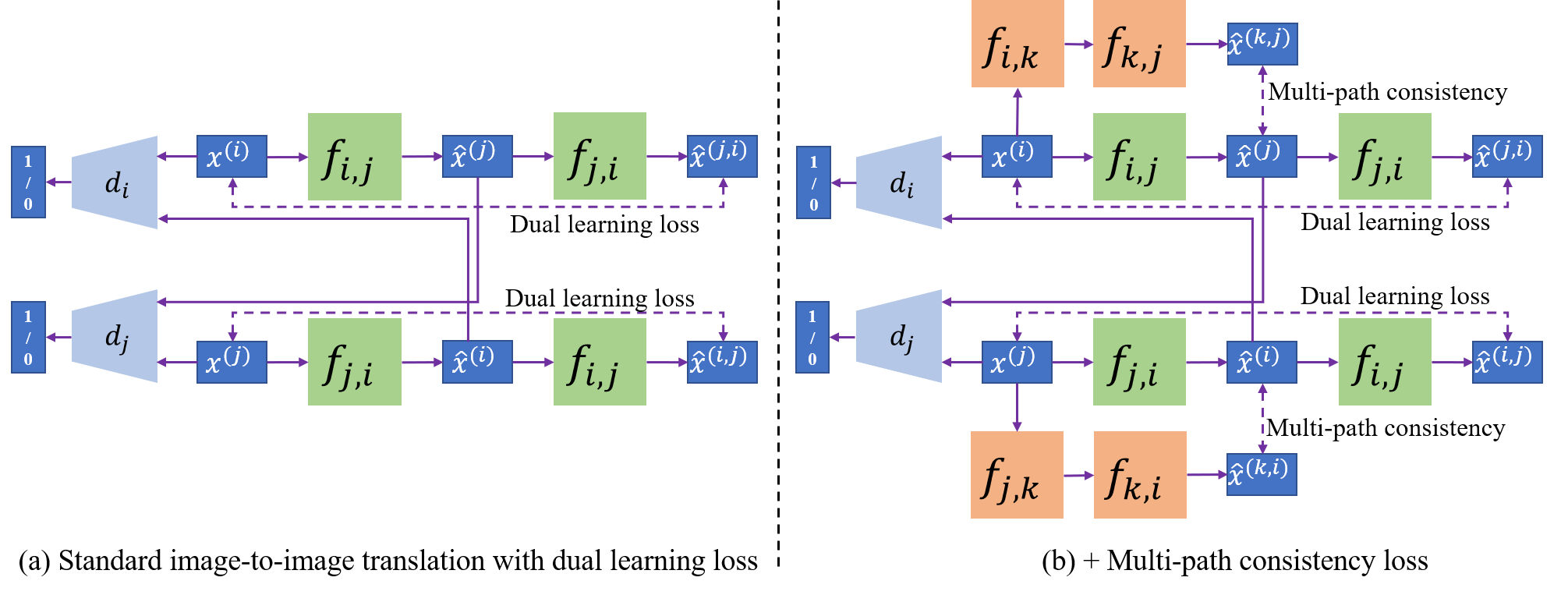}
\caption{The standard and our proposed frameworks of image-to-image translation, where $\hat{x}^{(j)}=f_{i,j}(x^{(i)})$, $\hat{x}^{(i)}=f_{j,i}(x^{(j)})$, $\hat{x}^{(j,i)}=f_{j,i}(\hat{x}^{(j)})$,
$\hat{x}^{(i,j)}=f_{i,j}(\hat{x}^{(i)})$,
$\hat{x}^{(k,j)}=f_{k,j}(f_{i,k}(x^{(i)}))$,  $\hat{x}^{(k,i)}=f_{k,i}(f_{j,k}(x^{(j)}))$ .}
\label{fig:multi_path_loss}
\end{figure*}
\subsection{Translation between $\mathcal{D}_i$ and $\mathcal{D}_j$ with an auxiliary domain $\mathcal{D}_k$}\label{sec:i-jwk}
To effectively obtain the two translation models $f_{i,j}$ and $f_{j,i}$ with an auxiliary domain $\mathcal{D}_k$, we need the following additional components in the system: (1) Three discriminators $d_i$, $d_j$ and $d_k$, which are used to classify whether an input image is a natural one or an image generated by the machine. Mathematically, $d_l:\mathcal{D}_l\mapsto [0,1]$ models the probability that the input image is a natural image in domain $\mathcal{D}_l$, $l\in\{i,j,k\}$. (2) Four auxiliary mappings $f_{i,k}$, $f_{k,i}$, $f_{j,k}$ and $f_{k,j}$, which are all related to $\mathcal{D}_k$.

Considering that deep learning algorithms usually iteratively work on mini-batches of data instead of the whole training datasets at the same time, in the remaining part of this section, we describe how the models are updated on the batches $\mathcal{B}_l\subseteq\mathcal{D}_l$, $l\in\{i,j,k\}$, where $\mathcal{B}_l$ is a minibatch of data in  $\mathcal{D}_l$.

The training loss consists of three parts:

\noindent(1) {\em Dual learning loss} between $\mathcal{D}_i$ and $\mathcal{D}_j$, which models the reconstruction duality between $f_{i,j}$ and $f_{j,i}$. Mathematically,
\begin{equation}
\begin{aligned}
\ell^{i,j}_{d}=&\frac{1}{\vert\mathcal{B}_i\vert}\sum_{x^{(i)}\in\mathcal{B}_i}\Vert x^{(i)}-f_{j,i}(f_{i,j}(x^{(i)}))\Vert_1 +\\
&\frac{1}{\vert\mathcal{B}_j\vert}\sum_{x^{(j)}\in\mathcal{B}_j}\Vert x^{(j)}-f_{i,j}(f_{j,i}(x^{(j)}))\Vert_1,
\end{aligned}
\label{eq:dual_loss}
\end{equation}
where $\vert\mathcal{B}_i\vert$ and $\vert\mathcal{B}_j\vert$  are the numbers of images in mini-batch $\mathcal{B}_i$ and mini-batch $\mathcal{B}_j$.

\noindent(2) {\em Multi-path consistency loss} with an auxiliary domain $\mathcal{D}_k$, which regularizes the training by leveraging the information provided by the third domain.  Mathematically,
\begin{equation}
\begin{aligned}
\ell^{i,j|k}_{c}=&\frac{1}{\vert\mathcal{B}_i\vert}\sum_{x^{(i)}\in\mathcal{B}_i}\Vert f_{i,j}(x^{(i)})-f_{k,j}(f_{i,k}(x^{(i)}))\Vert_1 +\\
&\frac{1}{\vert\mathcal{B}_j\vert}\sum_{x^{(j)}\in\mathcal{B}_j}\Vert f_{j,i}(x^{(j)})-f_{k,i}(f_{j,k}(x^{(j)}))\Vert_1.
\end{aligned}
\label{eq:consist_loss}
\end{equation}
\noindent(3) {\em GAN loss}, which enforces the generated images to be natural enough.  Let $\hat{\mathcal{B}}_l$ denote a collection of all generated/fake images to domain $\mathcal{D}_l$. When $l=k$, $\hat{\mathcal{B}}_l=\{f_{i,k}(x^{(i)})|x^{(i)}\in\mathcal{B}_i  \}\cup\{f_{j,k}(x^{(j)})|x^{(j)}\in\mathcal{B}_j  \}$. When $l\in\{i,j\}$, $\hat{\mathcal{B}}_l$ is the combination of one-hop translation and two-hop translation, defined as $\{f_{p,l}(x^{(p)})|x^{(p)}\in\mathcal{B}_p  \}\cup\{f_{k,l}(f_{p,k}(x^{(p)}))|x^{(p)}\in\mathcal{B}_p\}$ where $p=\{i,j\}\backslash\{l\}$. The GAN loss is defined as follows:
\begin{equation}
\begin{aligned}
\ell^{i,j|k}_{\text{GAN}}=\sum_{l\in\{i,j,k\}}\Big\{&\frac{1}{\vert\mathcal{B}_l\vert}\sum_{x^{(l)}\in\mathcal{B}_1}[\log d_l(x^{(l)})] \\
+&\frac{1}{\vert\hat{\mathcal{B}}_l\vert}\sum_{\hat{x}^{(l)}\in\hat{\mathcal{B}}_l}[\log (1-d_l(\hat{x}^{(l)}))]\Big\}.
\end{aligned}
\label{eq:gan_loss}
\end{equation}
All $f_{\cdot,\cdot}$'s work together to minimize the GAN loss, while all $d_\cdot$'s try to enlarge the GAN loss.

Given the aforementioned three kinds of loss, the overall loss can be defined as follows:
\begin{equation}
\ell_{\text{total}}^{i,j|k}(\mathcal{B}_i,\mathcal{B}_j,\mathcal{B}_k)= \ell^{i,j}_{d} + \ell^{i,j|k}_{c} + \alpha\ell^{i,j|k}_{\text{GAN}},
\label{eq:total_loss}
\end{equation}
where $\alpha$ is a hyper-parameter balancing the tradeoff between the GAN loss and other losses. All the six generators $f_{\cdot,\cdot}$'s work on minimizing Eqn.~\eqref{eq:total_loss} while all three discriminators $d_\cdot$'s work on maximizing Eqn.~\eqref{eq:total_loss}.

\subsection{Multi-domain image translation}
For an $N$-domain translation problem, when $N\ge3$, at each training iteration, it is too costly to build the consistency loss for each three domains. Alternatively, we can randomly select three domains $\mathcal{D}_i$, $\mathcal{D}_j$, $\mathcal{D}_k$ and build the consistency loss as follows:
\begin{equation}
\begin{aligned}
&\ell^{i,j,k}_{\text{total}}(\mathcal{B}_i,\mathcal{B}_j,\mathcal{B}_k) = \ell^{i,j|k}_{\text{total}}(\mathcal{B}_i,\mathcal{B}_j,\mathcal{B}_k) \\
+&\ell^{i,k|j}_{\text{total}}(\mathcal{B}_i,\mathcal{B}_k,\mathcal{B}_j) + \ell^{j,k|i}_{\text{total}}(\mathcal{B}_j,\mathcal{B}_k,\mathcal{B}_i),
\end{aligned}
\label{eq:overall_loss_gold}
\end{equation}
where the $\ell^{i,j|k}_{\text{total}}$ is defined in Eqn.~\eqref{eq:total_loss} and the other notations can be similarly defined.

\noindent{\bf Discussion}

\noindent(1) When $N=2$, we need to find the third domain as a auxiliary domain to help establish consistency. In this case, we can use Eqn.~\eqref{eq:total_loss} as the training objective, without applying consistency loss on the third domain. We work on the de-raining and de-noising tasks to verify such a case. (See Section~\ref{sec:two_domain_exps}.)

\noindent(2) We can build the consistency loss with  longer paths, e.g., the translation from $\mathcal{D}_1\to\mathcal{D}_3\cdots\mathcal{D}_N\to\mathcal{D}_2$ should be consistent with $\mathcal{D}_1\to\mathcal{D}_2$. Considering computation resource limitation, we leave this study to future work.

\subsection{Connection with StarGAN}
For an $N$-domain translation, when $N$ is large, it is impractical to learn $N(N-1)$ mappings. StarGAN~\cite{choi2017stargan} is a recently proposed method which uses a single model with different target labels to achieve image translation. With StarGAN, the mapping from $\mathcal{D}_i$ to $\mathcal{D}_j$ could be specified as $f(x^{(i)},c_j)$ where $x^{(i)}\in\mathcal{D}_i$, $f$ is shared among all tasks and $c_j$ is a learnable vector used to identify $\mathcal{D}_j$. All the generators share a same copy of the parameters except the target domain labels.

In terms of the discriminator, StarGAN only consists of one network which is not only used for justifying whether an image is a natural or fake one, but also serving as a classifier that distinguishes which domain does the input belong to. We also adopt such a kind of discriminator when using StarGAN as the basic model architecture. In this case, let $d_{\text{cls}}(l|x)$ denote the probability that the input $x$ is categorized as an image from domain $l$, $l\in[N]$. Following the notations in Section~\ref{sec:i-jwk}, the classification cost $\ell^{i,j|k}_{\text{cls},r}$ of real images and fake images $\ell^{i,j|k}_{\text{cls},f}$ for StarGAN can be formulated as follows:
\begin{equation}
\begin{aligned}
\ell^{i,j|k}_{\text{cls},r}=&\sum_{l\in\{i,j,k\}}\frac{-1}{\vert\mathcal{B}_l\vert}\sum_{x^{(l)}\in\mathcal{B}_1}[\log d_{\text{cls}}(l|x^{(l)})], \\
\ell^{i,j|k}_{\text{cls},f}=&\sum_{l\in\{i,j,k\}}
\frac{-1}{\vert\hat{\mathcal{B}}_l\vert}\sum_{\hat{x}^{(l)}\in\hat{\mathcal{B}}_l}[\log d_{\text{cls}}(l|\hat{x}^{(l)})].
\end{aligned}
\label{eq:cls_loss}
\end{equation}
When using StarGAN with the aforementioned classification errors, the image generators and discriminators cannot share a common objective function. Therefore, the loss function with multi-path consistency regularization, i.e., Eqn.~\eqref{eq:total_loss}, should be split and re-formulated as follows:
\begin{equation}
\begin{aligned}
\ell_{\text{total},G}^{i,j|k}(\mathcal{B}_i,\mathcal{B}_j,\mathcal{B}_k)&= \ell^{i,j}_{d} + \ell^{i,j|k}_{c} + \alpha\ell^{i,j|k}_{\text{GAN}}+\beta\ell^{i,j|k}_{\text{cls},f},\\
\ell_{\text{total},D}^{i,j|k}(\mathcal{B}_i,\mathcal{B}_j,\mathcal{B}_k)&= - \alpha\ell^{i,j|k}_{\text{GAN}}+\beta\ell^{i,j|k}_{\text{cls},r},
\label{eq:total_loss_v2}
\end{aligned}
\end{equation}
where both $\alpha$ and $\beta$ are the hyper-parameters. The generator and discriminator should try to minimize $\ell_{\text{total},G}^{i,j|k}$ and $\ell_{\text{total},D}^{i,j|k}$ respectively. Also, Eqn~\eqref{eq:overall_loss_gold} should be re-defined accordingly.

\section{Experiments on multi-domain translation}\label{experiment}
For multi-domain translation, we carry out two groups of experiments to verify our proposed framework, which are face-to-face translation with different attributes and paint-to-photo translation with different art styles. We choose StarGAN~\cite{choi2017stargan}, a state-of-the-art algorithm on multi-domain image translation as our baseline.

\subsection{Setting}\label{implement}
\noindent{\bf Datasets} For multi-domain face-to-face translation, we use the CelebA dataset~\cite{liu2015faceattributes}, which consists of $202,599$ face images of celebrities. Following~\cite{choi2017stargan}, we select seven attributes and build seven domains correspondingly. Among these attributes, three of them represent hair color, including black hair, blond hair, brown hair; two of them represent the gender, i.e., male and female; the left two represent age, including old and young. Note that these seven features are not disjoint. That is, a man can both have blond hair and be young.

For multi-domain paint-to-photo translation, we use the paintings and photographs collected by \cite{Zhu_2017_ICCV}, where we construct five domains including Cezanne, Monet, Ukiyo-e, Vangogh and photographs.

\noindent{\bf Architecture} For multi-domain translation tasks, we choose StarGAN~\cite{choi2017stargan} as our basic structure. One reason is that for an $N$-domain translation, we need $N(N-1)$ independent models to achieve translations between any two domains; with StarGAN, we only need one model. Another reason is that \cite{choi2017stargan} claim that on face-to-face translation, StarGAN achieves better performance for multi-domain translation compared to simply using multiple CycleGANs since multi-tasks are involved in the same model and the common knowledge among different tasks can be shared to achieve better performance.

\noindent{\bf Optimization} We use Adam optimizer~\cite{kingma2014adam} with learning rate $0.0001$ for the first $10$ epochs and linearly decay the learning rate every $10$ epochs. All the models are trained on one NVIDIA K40 GPU for one day. The $\alpha$ in Eqn.~\eqref{eq:total_loss} and Eqn.~\eqref{eq:total_loss_v2} is set to $0.1$, and $\beta$ in Eqn.~\eqref{eq:total_loss_v2} is also set to $0.1$.

\noindent{\bf Evaluation} We take both qualitative and quantitative analysis to verify the experiment results. For qualitative analysis, we visualize the results of both the baseline and our algorithm, and compare their differences. For quantitative analysis, following~\cite{choi2017stargan}, we perform classification experiments on generation synthesis. We train the classifiers on the image-translation training data using the same architecture as that for the discriminator, resulting in near-perfect accuracies, and then compute the classification error rates of generated images based on the classifiers. The Fréchet Inception Distance (FID) \cite{NIPS2017_7240} that measures similarity between generated image dataset and real image dataset is used to evaluate translated results quality. The lower the FID is, the better the translation results are. We also carry out user study for the generated images.

\begin{figure}[!htb]
\centering
\includegraphics[width=1\linewidth]{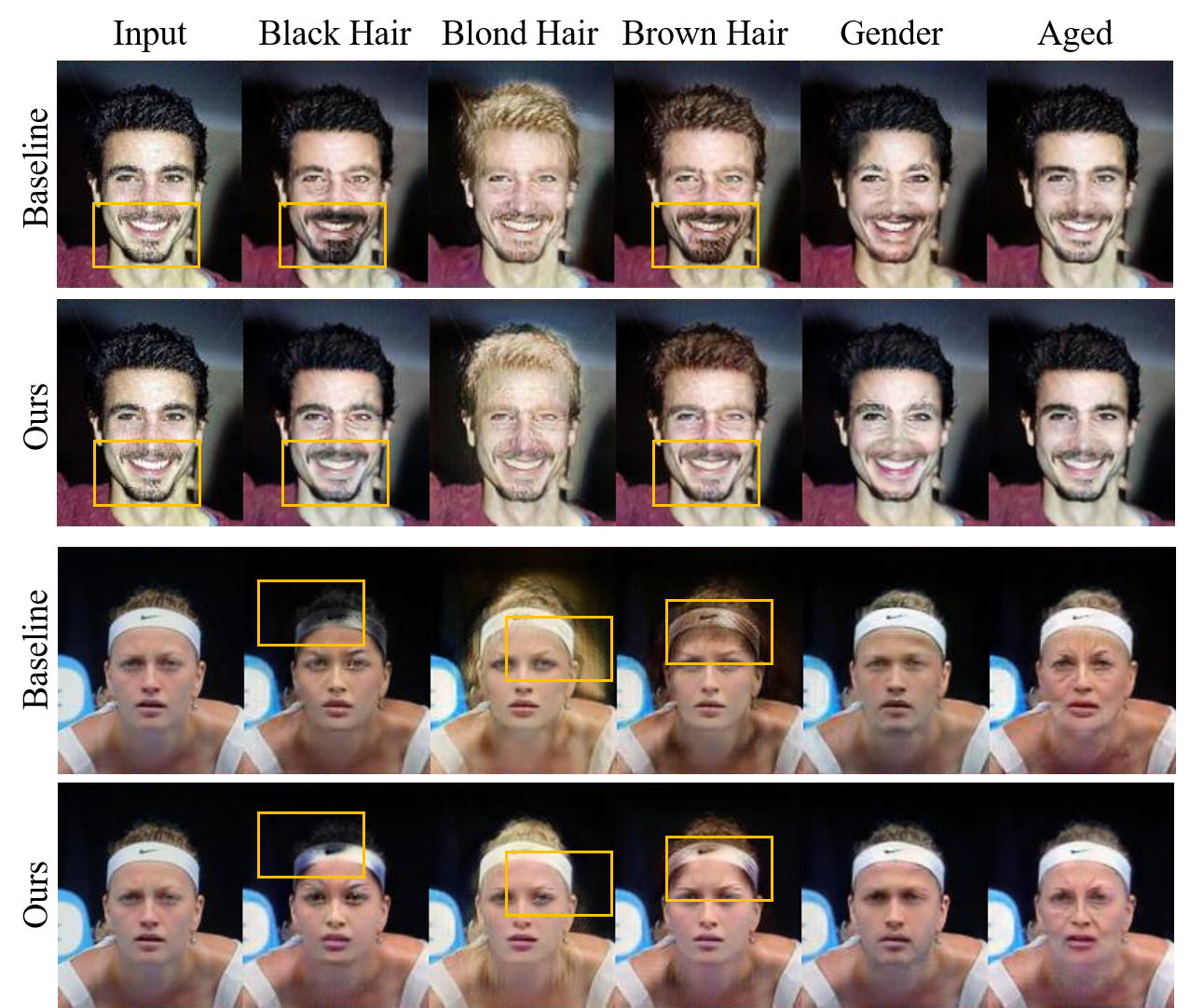}
\caption{Two groups of multi-domain face-to-face translation results. The rows started with ``Baseline" and ``Ours" represent the results of the baseline and our method.}
\label{fig:face-to-face}
\end{figure}
\begin{figure*}[!htb]
\centering
\includegraphics[width=1\linewidth]{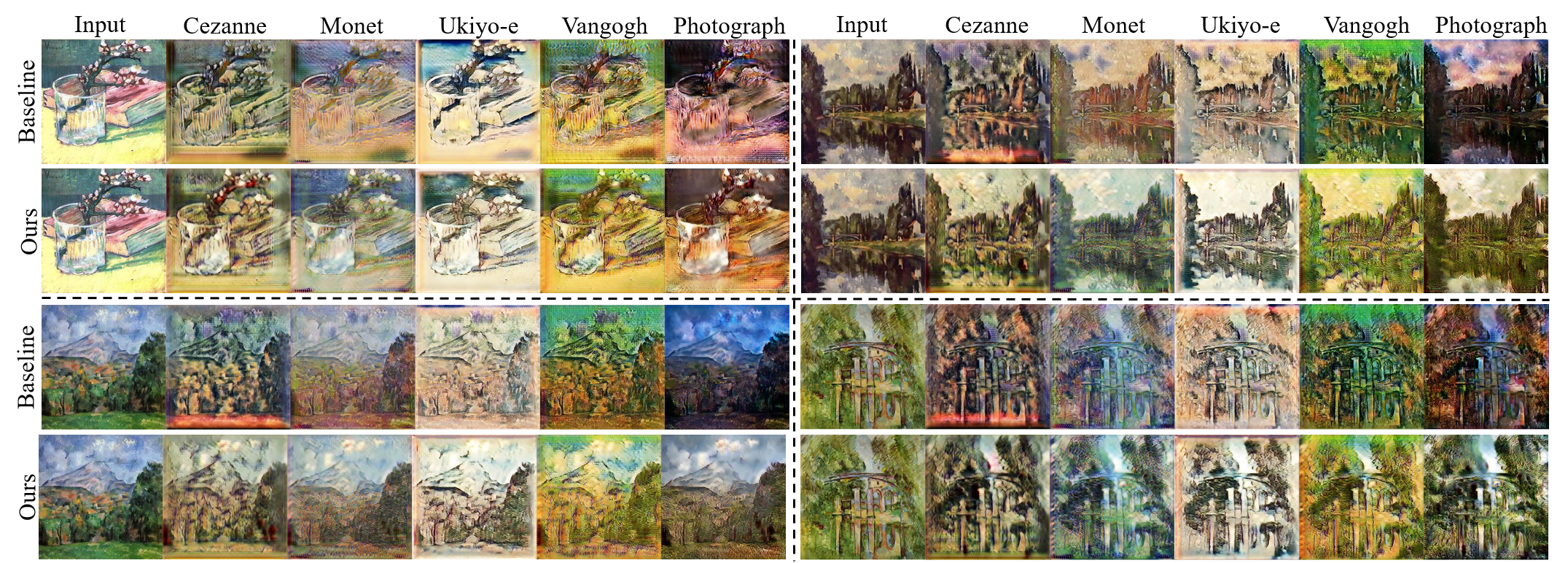}
\caption{ Multi-domain paint-to-photo translation results. }
\label{fig:paint-to-photo}
\end{figure*}
\subsection{Face-to-face translation}
The results of face-to-face translation are shown in Figure~\ref{fig:face-to-face}. In general, both the baseline and our proposed method can successfully transfer the images to the target domain. But there are many places where our proposed method outperforms the baseline:

\noindent(1) Our proposed method could preserve more information of the input images. Take the translations in the upper part as an example. To translate the input to black hair and blond hair domain, the domain-specific feature \cite{lin2018conditional} to be changed is the hair color, while the other domain-independent features should be kept as many as possible. The baseline algorithm changes the beard of input images, while our algorithm keeps such a feature due to the multi-path consistency regularization. Since a consistency regularization requires both the one-hop translation and two-hop translation to preserve enough similarity, and the two-hop translation path is independent of one-hop translation path, errors/distortions in the translation results are more likely to be avoided from the aspect of probability. As a result, our proposed method can carry more information of the original image.

\noindent(2) After applying multi-path consistency regularization, our algorithm could generate clearer images with less noise than the baseline, like the images with black hair and blond hair in the bottom part. One possible reason is that multi-path consistency pushes the models to generate consist images. The random noise would affect the consistency and our proposed regularization way could reduce such effects.


For quantitative evaluation, the results are shown in Table~\ref{table:face-to-face} and Table~\ref{table:face-to-face_fid}.
We can see that our proposed method achieves lower classification error rates on three different domains,  which demonstrates that our generator could produce images with more significant features in the target domain. For FID score, our algorithm also achieves $1.79$ improvement, which suggests that our translation results's distribution are more similar to real ones'.

\begin{table}[htb!]
\centering
\begin{tabular}{c c c c}
\toprule
&Hair Color &Gender &Age \\
\midrule
Baseline &19.01\%  & 11.60\%& 25.52\%\\
Ours   & 17.08\% & 10.23\%&24.39 \% \\
\midrule
 Improvements & 1.93\% & 1.37\%&1.13\%\\
\bottomrule
\end{tabular}
\caption{Classification error rates of face-to-face translation.}
\label{table:face-to-face}
\end{table}

\begin{table}[htb!]
\centering
\begin{tabular}{c c c}
\toprule
Baseline & Ours  &Improvement\\
\midrule
20.15 & 18.36 &1.79\\
\bottomrule
\end{tabular}
\caption{FID scores of face-to-face translation.}
\label{table:face-to-face_fid}
\end{table}

\subsection{Paint-to-photo translation}
The results of multi-domain paint-to-photo translation are shown in Figure~\ref{fig:paint-to-photo}. Again, the models trained with multi-path consistency loss outperform the baselines. For example, our model can generate more domain-specific paintings than the baseline method as shown in the generated Vangogh paintings. We also observe that our model effectively reduces the block artifact in the translation results, such as generated Cezanne, Monet and Vangogh paintings. Besides, our model prefers generating images with clearer edges and context. As shown in the upper-left corner of Figure~\ref{fig:paint-to-photo}, we could generate images with obvious edges and content, while the baseline algorithm fails with unclear and messy generations.

Similar to face-to-face translation, we also show the classification errors and FID. The results are in Table~\ref{table:paint-to-photo} and Table~\ref{table:paint-to-photo_fid}. Our algorithm achieves significantly better results than the baseline, which demonstrates the effectiveness of our method.

\begin{table}[htb!]
\centering
\begin{tabular}{c c c}
\toprule
Baseline & Ours  &Improvement\\
\midrule
35.52\% & 30.17\% &5.35\%\\
\bottomrule
\end{tabular}
\caption{Classification error rates of paint-to-photo translation.}
\label{table:paint-to-photo}
\end{table}

\begin{table}[htb!]
\centering
\scalebox{0.75}{
\begin{tabular}{c c c c c c}
\toprule
&Cezanne &Monet &Ukiyo-e & Vangogh &Photograph \\
\midrule
Baseline & 219.43 &199.77 &163.46 & 226.77&79.33\\
Ours   & 210.82 & 170.52&154.29 & 216.78&64.10\\
\midrule
 Improvements & 8.61 &29.25 & 9.17& 9.99&15.23\\
\bottomrule
\end{tabular}}
\caption{FID scores of paint-to-photo translation.}
\label{table:paint-to-photo_fid}
\end{table}
\begin{figure*}[!htb]
\centering
\includegraphics[width=1\linewidth]{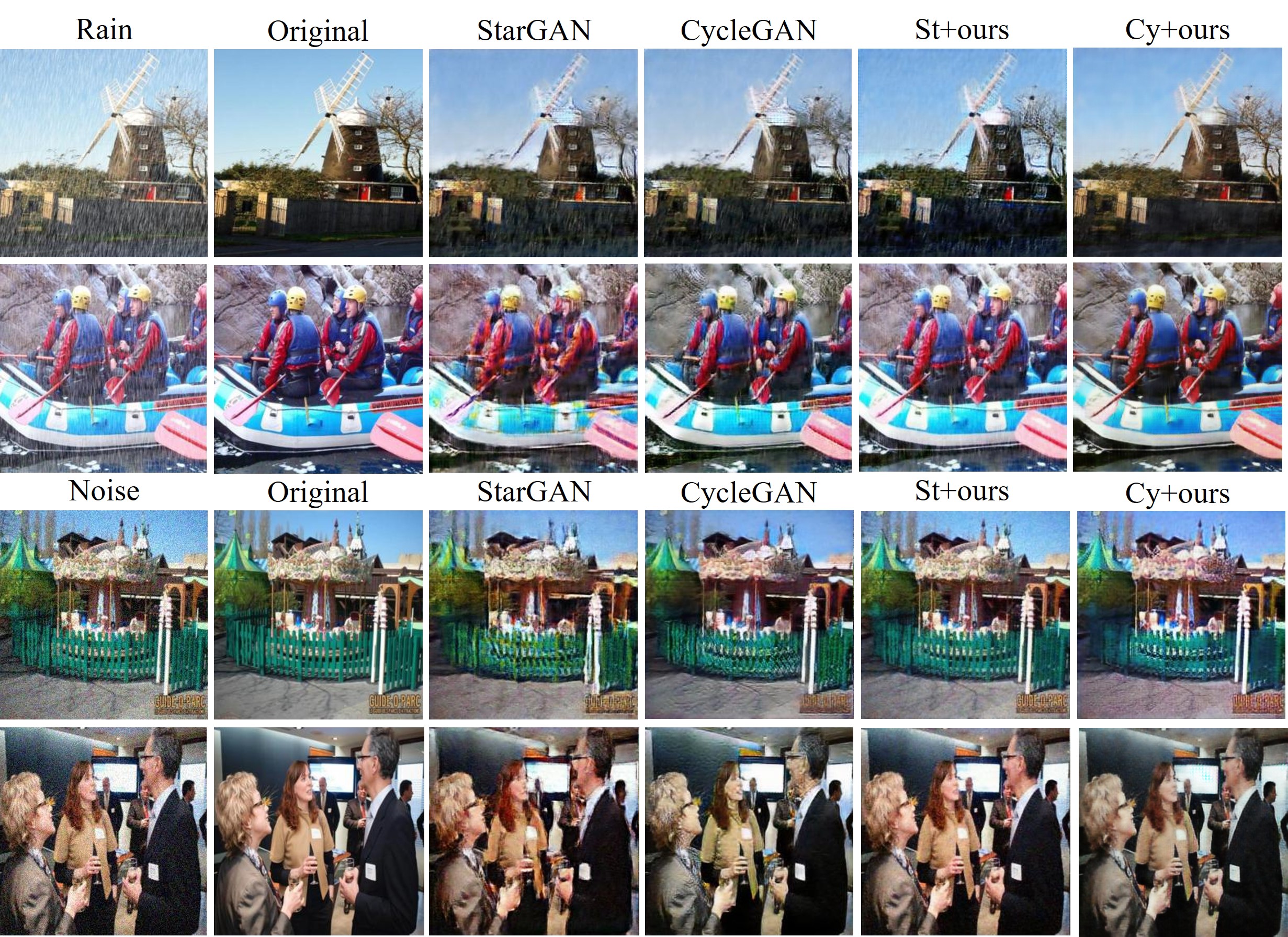}
\caption{Unsupervised de-raining (first two rows) and de-noising (last two rows) results. From left to right, the columns represent the rainy/noisy input, the original clean image, the results of StarGAN (St) and CycleGAN (Cy) without multi-path consistency loss, and the corresponding results with our method (St+ours, Cy+ours) respectively.}
\label{fig:noise-to-photo}
\end{figure*}
\subsection{User study}
We carry out user study to further evaluate our results. 20 users with diverse education backgrounds are chosen as reviewers. We randomly select $35$ groups of generated images for the face-to-face and paint-to-photo translation, where each group contains the translation results from the same input image to different categories of both the baseline and our algorithm. For any group of experiments, reviewers have to choose the better one without knowing which algorithm the images are generated from.

The statistics are shown in Table~\ref{table:usr_study}. Among the $700$ votes for face-to-face translation, $59.14\%$ belongs to our proposed algorithm, and for paint-to-photo, $89.85\%$ belongs to ours. The user study shows that we achieve better performance than the baseline, especially for paint-to-photo translation.
\begin{table}[!htb]
\centering
\begin{tabular}{cccc}
\toprule
\multicolumn{2}{c}{face-to-face} & \multicolumn{2}{c}{paint-to-photo}  \\
\midrule
Baseline & Ours & Baseline & Ours \\
\midrule
40.86\% & 59.14\% & 10.14\% & 89.85\% \\
\bottomrule
\end{tabular}
\caption{Statistics of user study.}
\label{table:usr_study}
\end{table}

\section{Experiments on two-domain translation}\label{sec:two_domain_exps}
In this section, we work on two-domain translations with auxiliary domains. We choose two different tasks, unsupervised de-raining and unsupervised de-noising, which means to remove the rain or noise from the input images.
\subsection{Setting}
\noindent{\bf Datasets} We use the raining images and original images collected by \cite{fu2017removing,yang2017deep}. For unsupervised translation, we randomly shuffle the raining images and the original ones to obtain an unaligned dataset. As for the unaligned dataset for de-noising, we add uniform noise to each original image and then shuffle them. For the de-raining and de-noising experiments, we choose the noise image domain and raining image domain as auxiliary ones respectively.

\noindent{\bf Architecture} We first choose StarGAN as the basic network architecture. The model architecture is the same as that used in Section~\ref{implement}. In addition, to verify the generality of our framework, we also apply CycleGAN~\cite{Zhu_2017_ICCV} to this task. To combine with our framework, a total of six generators and three discriminators are implemented. We follow Section~\ref{sec:i-jwk} to jointly optimize the de-rain and add-rain networks for the de-raining task with consistency loss built upon the domains with images of random noise. Similar method is also applied to the de-noising task.

\noindent{\bf Evaluation} For qualitative analysis, again we compare the images generated by the baseline and our algorithm. For quantitative analysis, except for the classification errors, we check the Peak Signal-to-Noise Ratio (briefly, PSNR), of the generated images with the original images. The larger PSNR is, the better the restoration quality is.

\subsection{Results}
The unsupervised de-raining and unsupervised de-noising results are shown in Figure~\ref{fig:noise-to-photo}. On the two tasks, our proposed method can improve both StarGAN and CycleGAN and generate cleaner images with less block artifacts, smoother colors and clearer facial expressions. We also find that in de-raining and de-noising tasks, CycleGAN outperforms StarGAN and can generate images with less rain and noise. One reason is that unlike face-to-face translation whose domain-independent features are centralized and easy to capture, natural scenes are usually diverse and complex, in which a single StarGAN might not have enough capacity to model. In comparison, a CycleGAN works for two-direction translation only, which has enough capacity to model and de-rain/de-noise the images.

\begin{figure}[htb!]
\centering
\includegraphics[width=0.95\linewidth]{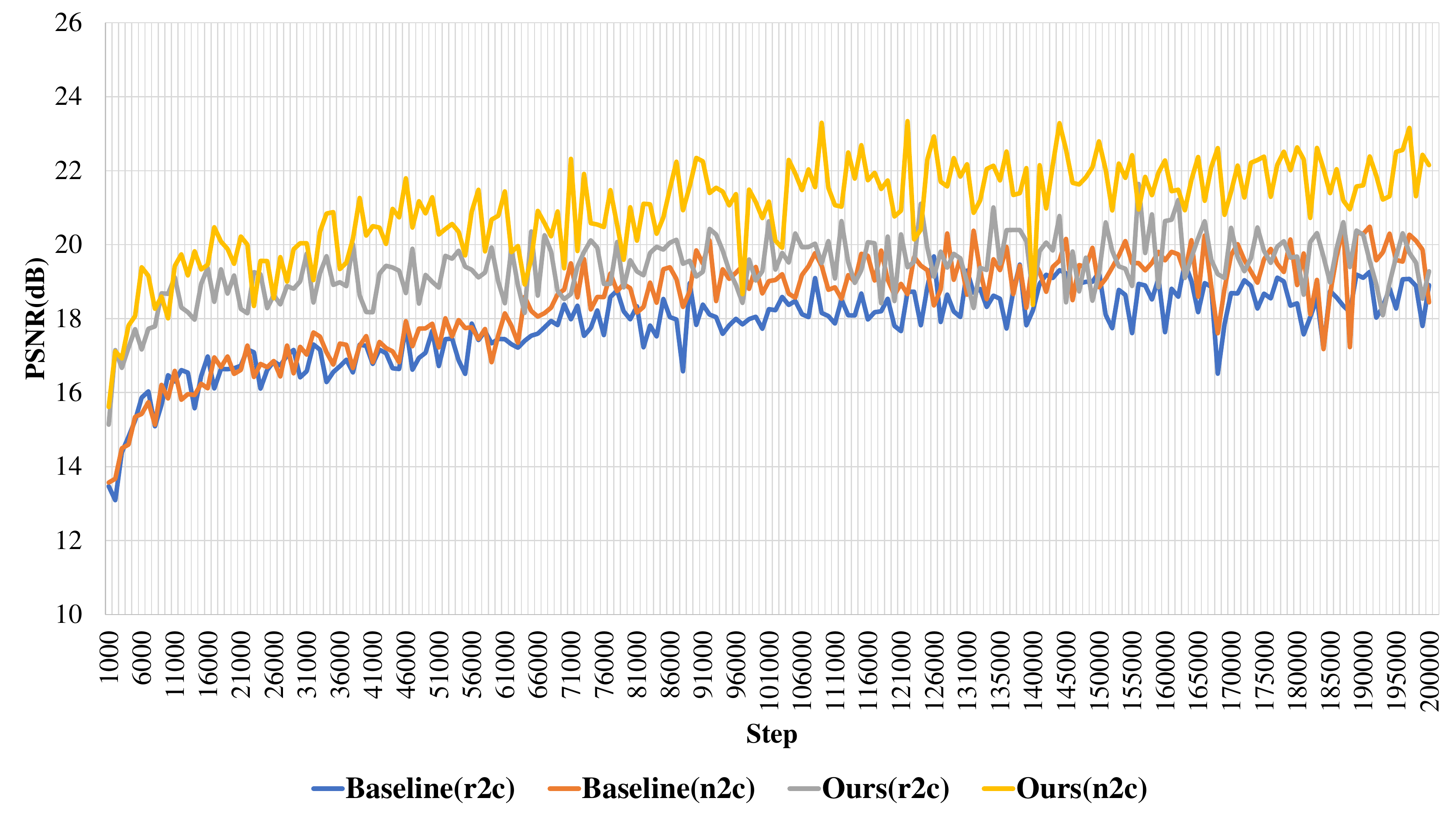}
\caption{ PSNR curve w.r.t training steps.}
\label{fig:psnr_curve}
\end{figure}

We report the classification error rates and PSNR (dB) of de-raining and de-noising in Table~\ref{table:noise-to-photo}. The classification error rates of StarGAN and CycleGAN before using multi-path consistency loss are $2.91\%$ and $1.93\%$ respectively, while after applying multi-path consistency loss the numbers are $1.70\%$ and $1.65\%$, which shows the efficiency of our method. In terms of PSNR, as shown in Table~\ref{table:noise-to-photo}, our method achieves higher scores, which means that our model has better restoration abilities. That is, for the two-domain translation, our framework still works. We also plot PSNR curves of the StarGAN based models on the test set w.r.t. training steps. The results are shown in Figure~\ref{fig:psnr_curve}. On the two tasks, training with multi-path consistency regularization could always achieve higher PSNR than the corresponding baseline. This shows that our proposed method can achieve not only higher PSNR values, but also faster convergence speed.

\begin{table}[htb!]
	\footnotesize
	\setlength{\abovecaptionskip}{0.cm}
	\setlength{\belowcaptionskip}{-0.cm}
	\centering
	\scalebox{1.0}{
		\begin{tabular}{c c c c}
			\toprule
			Method    & Classification Error & r$\rightarrow$c (dB) &n$\rightarrow$c (dB)  \\
			\midrule
			StarGAN       & 2.91\% & 19.43 &20.39\\
            CycleGAN       & 1.93\% & 20.87 &21.99\\
			Ours(St)    & 1.70\% & 21.13 &23.25\\
            Ours(Cy)    & 1.65\% & 21.21 &23.28\\
			
			\bottomrule
	\end{tabular}}
    	\caption{Classification error rates and PSNR (dB) of de-raining and de-noising translation results.}
    \label{table:noise-to-photo}
\end{table}

\section{Conclusion and future work}\label{conclusion}
In this paper, we propose a new kind of loss, multi-path consistency loss, which can leverage the information of multiple domains to regularize the training. We provide an effective way to optimize such a framework under multi-domain translation environments. Qualitative and quantitative results on multiple tasks demonstrate the effectiveness of our method.

For future work, it is worth studying what would happen if more than three paths are included. In addition, we will generalize multi-path consistency by using stochastic latent variables as the auxiliary domain
\section{Acknowledgement}
This work was supported in part by NSFC under Grant 61571413, 61632001.
{\small
\bibliographystyle{named}
\bibliography{Bibliography-File}
}

\end{document}